\documentclass{article}
\usepackage{spconf}
\usepackage{amsmath}
\usepackage{amssymb}
\usepackage{graphicx}
\usepackage{xspace}
\usepackage{xcolor}
\usepackage{bm}
\usepackage{hyperref}

\def\eg{\emph{e.g.}\xspace} 
\def\ie{\emph{i.e.}\xspace}

\def\wrt{w.r.t.\xspace}

\newcommand{\figref}[1]{{figure~\ref{#1}}}
\newcommand{\Figref}[1]{{Figure~\ref{#1}}}
\newcommand{\secref}[1]{{section~\ref{#1}}}

\renewcommand{\eqref}[1]{{equation~(\ref{#1})}}

\newcommand{\qt}[1]{`{#1}'}

\renewcommand{\vec}[1]{{\bf #1}}

\newcommand\comment[1]{}
\newcommand\commentresolved[1]{}

\usepackage{array}
\newcolumntype{L}[1]{>{\raggedright\let\newline\\\arraybackslash\hspace{0pt}}m{#1}}
\newcolumntype{C}[1]{>{\centering\let\newline\\\arraybackslash\hspace{0pt}}m{#1}}
\newcolumntype{R}[1]{>{\raggedleft\let\newline\\\arraybackslash\hspace{0pt}}m{#1}}
\newcolumntype{N}{@{}m{0pt}@{}}

\title{Domain adaptive segmentation in volume electron microscopy imaging}
\name{Joris Roels$^{1,3}$ \qquad Julian Hennies$^{4}$ \qquad Yvan Saeys$^{2,3}$ \qquad Wilfried Philips$^{1}$ \qquad Anna Kreshuk$^{4}$ \thanks{This research has been made possible by the Agency for Flanders Innovation \& Entrepreneurship (VLAIO). We gratefully acknowledge the support of NVIDIA Corporation with the donation of the Titan X Pascal GPU used for this research. We would like to thank Anna Steyer and Yannick Schwab (EMBL - Volume Correlative Light and Electron Microscopy) for the provided FIB-SEM HeLa dataset.}}

\address{
$^1$ Department of Telecommunications and Information Processing, \\Ghent University / IMEC, Ghent, Belgium \\ 
$^2$ Department of Applied Mathematics, Computer Science and Statistics, \\Ghent University, Ghent, Belgium \\
$^3$ Center for Inflammation Research, VIB, Ghent, Belgium \\
$^4$ Cell Biology and Biophysics, EMBL, Heidelberg, Germany
}
\begin{document}
%
\maketitle
\begin{abstract}
In the last years, automated segmentation has become a necessary tool for volume electron microscopy (EM) imaging. So far, the best performing techniques have been largely based on fully supervised encoder-decoder CNNs, requiring a substantial amount of annotated images. Domain Adaptation (DA) aims to alleviate the annotation burden by \qt{adapting} the networks trained on existing groundtruth data (source domain) to work on a different (target) domain with as little additional annotation as possible. Most DA research is focused on the classification task, whereas volume EM segmentation remains rather unexplored. In this work, we extend recently proposed classification DA techniques to an encoder-decoder layout and propose a novel method that adds a reconstruction decoder to the classical encoder-decoder segmentation in order to align source and target encoder features. The method has been validated on the task of segmenting mitochondria in EM volumes. We have performed DA from brain EM images to HeLa cells and from isotropic FIB/SEM volumes to anisotropic TEM volumes. In all cases, the proposed method has outperformed the extended classification DA techniques and the finetuning baseline. An implementation of our work can be found on \href{https://github.com/JorisRoels/domain-adaptive-segmentation}{https://github.com/JorisRoels/domain-adaptive-segmentation}. 
\end{abstract}
\section{Introduction}
\label{sec:introduction}
Recent developments in volume electron microscopy (EM) have dramatically increased the throughput and simplified the acquisition of large-scale datasets. The problem of segmenting the resulting volumes has also received a lot of attention \cite{Ronneberger2015,Chen2016,Oztel2017,Funke2018}. For a specific use-case (\eg segmentation of neuron circuits \cite{Ronneberger2015,Chen2016,Funke2018} or mitochondria \cite{Oztel2017}), the state-of-the-art workflows are based on training encoder-decoder networks using large amounts of pixel-level labels. The extracted features are typically data-dependent and high performance on slightly different datasets (\eg different microscope or sample preparation protocol) is therefore not always guaranteed. 

Domain adaptation (DA) tackles the problem of building a predictive model for a target dataset with no or very few labels by using a relatively large labeled source dataset. The state-of-the-art in (deep) DA is however largely focused on classification \cite{Ganin2016,Sun2016,Long2017} and an extension to segmentation is not straightforward. Recent developments in the field of segmentation show promising results \cite{Hoffman2016,Zhang2018}, even specifically for EM \cite{Bermudez-Chacon2018}. \commentresolved{Here another sentence is needed saying why these recent developments are not enough and what gap you are closing} However, they regularize only a small fraction of the extracted features or are based on adversarial networks, which are hard to optimize for end-users without significant deep learning expertise. In this work, we introduce a natural extension of classification-based DA techniques to encoder-decoder segmentation networks by regularizing the encoder features. This regularization significantly improves the network performance in the target domain over classical finetuning, at an additional computational cost. Furthermore, we propose a new unsupervised DA method for such networks based on auto-encoder feature alignment \cite{Ghifary2016,Hu2018} which avoids computationally intensive regularization metrics or challenging adversarial network training without sacrificing segmentation performance. 

\begin{figure*}[t!]
	\centering
    \includegraphics[width=\linewidth]{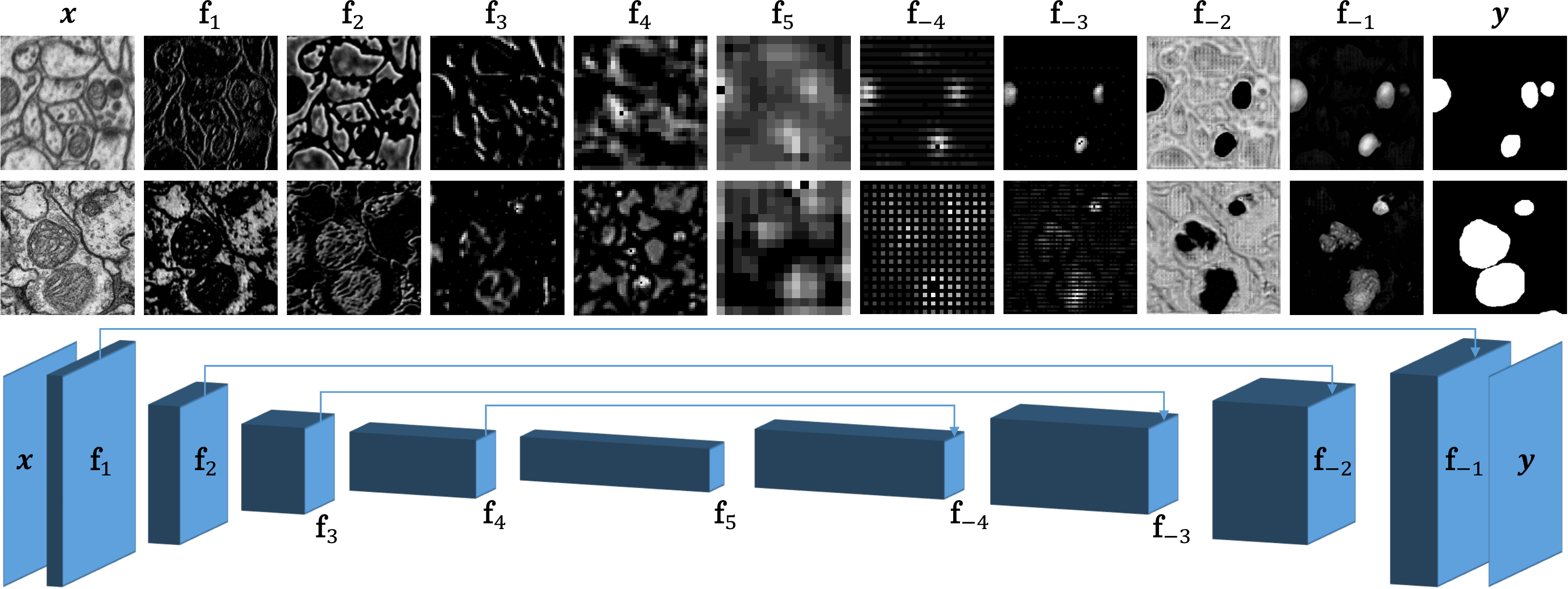} 
	\caption{Encoder-decoder segmentation network architecture with skip connections. The top and bottom row illustrate layer activations extracted from source and target data, respectively, with a network that was trained on the source. The domain shift is especially visible in the encoder features ($\vec{f}_i$), whereas the decoder features ($\vec{f}_{-i}$) are much closer to an actual segmentation result. This motivates discrepancy regularization on the encoder features. \commentresolved{temporary figure, actual data will be added}}
\label{fig:regularization}
\end{figure*}

We start with a brief overview of the related work in classification and segmentation DA (\secref{sec:related-work}). Next, we propose an extension of classification-based DA techniques to the segmentation task in \secref{sec:domain-adaptive-segmentation}. \secref{sec:y-net} describes the new unsupervised auto-encoder DA method in more detail (\secref{sec:y-net}). Finally, all methods are validated on two mitochondria segmentation use-cases in volume EM data (\secref{sec:results}). \commentresolved{I would put the refsecs in brackets}

\section{Related work}
\label{sec:related-work}

Segmentation in volume EM is a semantic segmentation problem where each pixel is to be assigned the appropriate class label. The current state-of-the-art is largely based on extracting features in the encoder through various convolution and pooling stages and returning to a segmentation at the original resolution through the decoder with skip-connections \cite{Ronneberger2015,Chen2016,Funke2018}. \commentresolved{shorten this: most work in segmentation just uses U-Net}

Most DA approaches are designed for classification and align source and target features by including a domain discrepancy loss. The work of \cite{Ganin2016} models this discrepancy by means of a distribution similarity metric termed maximum mean discrepancy (MMD). Alternatively, a feature correlation difference (CORAL) is proposed in \cite{Sun2016}. In \cite{Long2015}, domain classifiers and gradient reversal layers are introduced to align the feature distributions in an adversarial setup (DANN). \commentresolved{mention GAN approaches and motivate why we don't look at them further}

\commentresolved{The first sentence is redundant, you say this in the paragraph above and in the intro already. I would start this paragraph with the second sentence 'The first DA method...'} The first DA method for semantic segmentation was proposed in \cite{Hoffman2016}. It is based on classical CNN feature extractors where the last feature layer is aligned using an adversarial loss. The recent work of \cite{Zhang2018} also employs domain confusion for alignment, but additionally normalizes the visual appearance of source and target data. Alternatively, the work of \cite{Bermudez-Chacon2018} proposes shared decoders combined with MMD regularization on the final decoder activations. 

\commentresolved{Never say "similar to" or "based on" to describe your work, it makes it look incremental :)}

\section{Domain adaptation segmentation}
\label{sec:domain-adaptive-segmentation}
\commentresolved{clarify that this is where the novel stuff starts}
Inspired by \cite{Bermudez-Chacon2018}, we propose an extension of clas\-si\-fi\-ca\-tion-based DA to encoder-decoder segmentation, using the MMD, CORAL and DANN approaches. Unsupervised DA segmentation assumes a labeled source $\mathcal{S} = \{ (\vec{x}_i^s, \vec{y}_i^s) \}_{i=1,\dots,n^s}$ of images $\vec{x}_i^s \in \mathbb{R}^N$ and pixel-level labels $\vec{y}_i^s \in \{0,\dots,C-1\}^N$ and an unlabeled target $\mathcal{T} = \{ \vec{x}_i^t \}_{i=1,\dots,n^t}$ of images $\vec{x}_i^t \in \mathbb{R}^N$ for which the goal is to maximize target segmentation performance. In the semi-supervised setup, there is also a small amount of target labels $\vec{y}_i^t \in \{0,\dots,C-1\}^N$ available. For notational convenience, we define $\mathcal{L}_s$ as any segmentation loss (\eg cross entropy), $\hat{\vec{y}}^{s/t}$ is the output of the source/target segmentation network, $\vec{f}_i^{s/t}$ and $\vec{f}_{-i}^{s/t}$ are the final feature activations on level $i$ in respectively the encoder and decoder for source/target (see \figref{fig:regularization}). 

The discussed classification DA approaches (DANN, CORAL and MMD) include a domain regularization loss $\mathcal{L}_{d}$ on the source and target features. The aligned features are usually the final activations used for classification. In an encoder-decoder setup, the feature extractor and pixel-wise classifier are not that clearly separated due to the skip connections between encoder and decoder layers. Nevertheless, we denote that the encoder activations largely contain segmentation features, whereas the decoder activations largely serve for segmentation refining and resolution enhancement\commentresolved{Add figure with random activations from pretrained network} (see \figref{fig:regularization}). Additionally, high-resolution encoder features (\ie the first layers) require less alignment compared to the low-resolution (high-level) encoder features, which should be more domain-invariant. Therefore, we propose to regularize each encoder feature activation level in a weighted fashion, \ie: 
\begin{equation}\label{eq:regularization-loss}
\mathcal{L} = \mathcal{L}_{s}(\hat{\vec{y}}^s, \vec{y}^s) + \sum_i \lambda_i \mathcal{L}_{d}(\vec{f}_i^s, \vec{f}_i^t)
\end{equation}
where $\lambda_i$ are regularization parameters and increasing \wrt $i$. 

Note that a target segmentation loss can be added to the loss function in \eqref{eq:regularization-loss} in the semi-supervised case. However, we experienced that this limits the capacity of the segmentation network significantly. Therefore, the network is initially trained unsupervised and finetuned with the available target labels in the semi-supervised case. 

\section{Y-Net}
\label{sec:y-net}
\begin{figure}[t!]
	\centering
    \includegraphics[width=\linewidth]{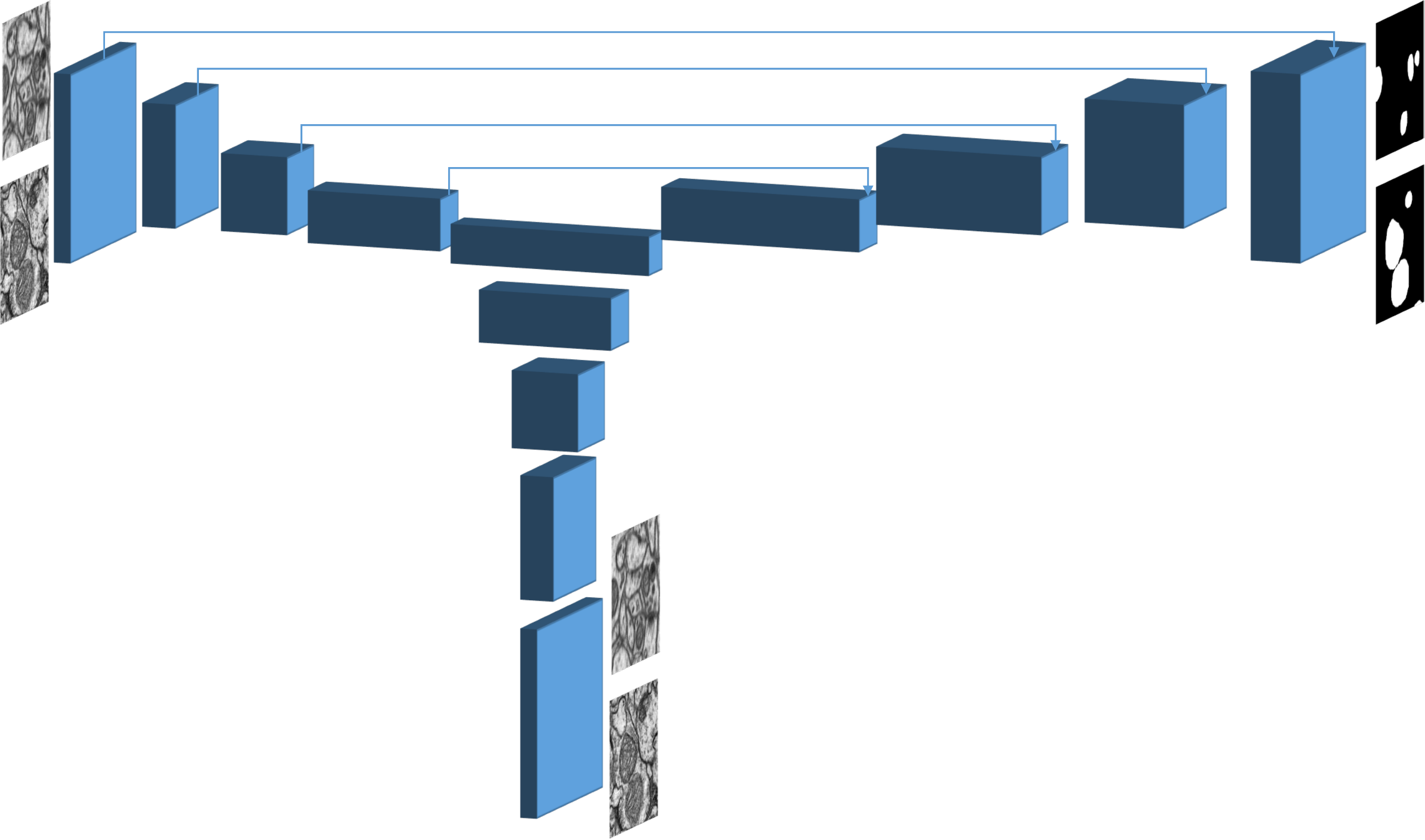} 
	\caption{Proposed unsupervised DA approach: a second decoder is attached to the encoder-decoder segmentation network which reconstructs both the source and target data.  \commentresolved{temporary figure, actual data will be added}}
\label{fig:ynet}
\end{figure}

\commentresolved{we need a transition from the previous section to this one}
The techniques discussed in the previous section compensate the domain shift between source and target domain by introducing feature (distribution) similarity metrics. They are, however, computationally intensive (\eg MMD computation, correlation matrix computation in CORAL, domain classification in DANN). The work of \cite{Ghifary2016}, however, shows that auto-encoders are able to extract generic useful features for classification, whereas the recent work of \cite{Hu2018} shows that these architectures are also able to align feature distributions. This motivates our idea of introducing a second decoder to the classical encoder-decoder setup which serves to reconstruct the input data which originates from both source and target domain (see \figref{fig:ynet}). The complete architecture is trained end-to-end with the following loss function: 
\begin{equation}
\mathcal{L} = \mathcal{L}_{s}(\hat{\vec{y}}^s, \vec{y}^s) + \lambda_r^s \mathcal{L}_{r}(\hat{\vec{x}}^s, \vec{x}^s) + \lambda_r^t \mathcal{L}_{r}(\hat{\vec{x}}^t, \vec{x}^t)
\end{equation}
where $\mathcal{L}_{r}$ is a reconstruction loss function (\eg mean-squared error), $\hat{\vec{x}}^{s/t}$ are reconstructions of the source/target inputs obtained by the auto-encoding sub-network and $\lambda_r^{s/t}$ are regularization parameters. 
The network is initially trained in an unsupervised fashion, after which the reconstruction decoder is discarded. Similar as in \secref{sec:domain-adaptive-segmentation}, the remaining segmentation network is finetuned on the target labels in the semi-supervised case. 

\section{Results \& discussion}
\label{sec:results}
We validate the discussed DA approaches on the problem of mitochondria segmentation in volume EM data. The source dataset consists of two annotated $165 \times 1024 \times 768$ FIB-SEM acquisitions (respectively for training and testing) of the CA1 hippocampus region at $5$ nm$^3$ isotropic resolution. We consider two target volumes. The first dataset (HeLa) consists of a $64 \times 512 \times 512$ annotated FIB-SEM block of a HeLa cell at $5$ nm lateral and $8$ nm axial resolution. The second dataset (Drosophila) \cite{Gerhard2013} is an annotated $20 \times 1024 \times 1024$ serial section Transmission Electron Microscopy (ssTEM) block of the Drosophila melanogaster third instar larva ventral nerve cord at $5$ nm lateral and $50$ nm axial resolution. \commentresolved{This should be more like 50 nm, it's fairly thick slices} Note that mitochondria in the HeLa data are significantly different from those in the source data and that the Drosophila data originates from a different modality and is highly anisotropic, which makes DA particularly challenging. \commentresolved{And one is acquired with very anisotropic resolution}. Both target datasets are split along the $x$ axis: $67\%$ and $33\%$ was used for training and testing, respectively. 

\begin{figure}[t!]
	\centering
    \includegraphics[width=0.48\linewidth]{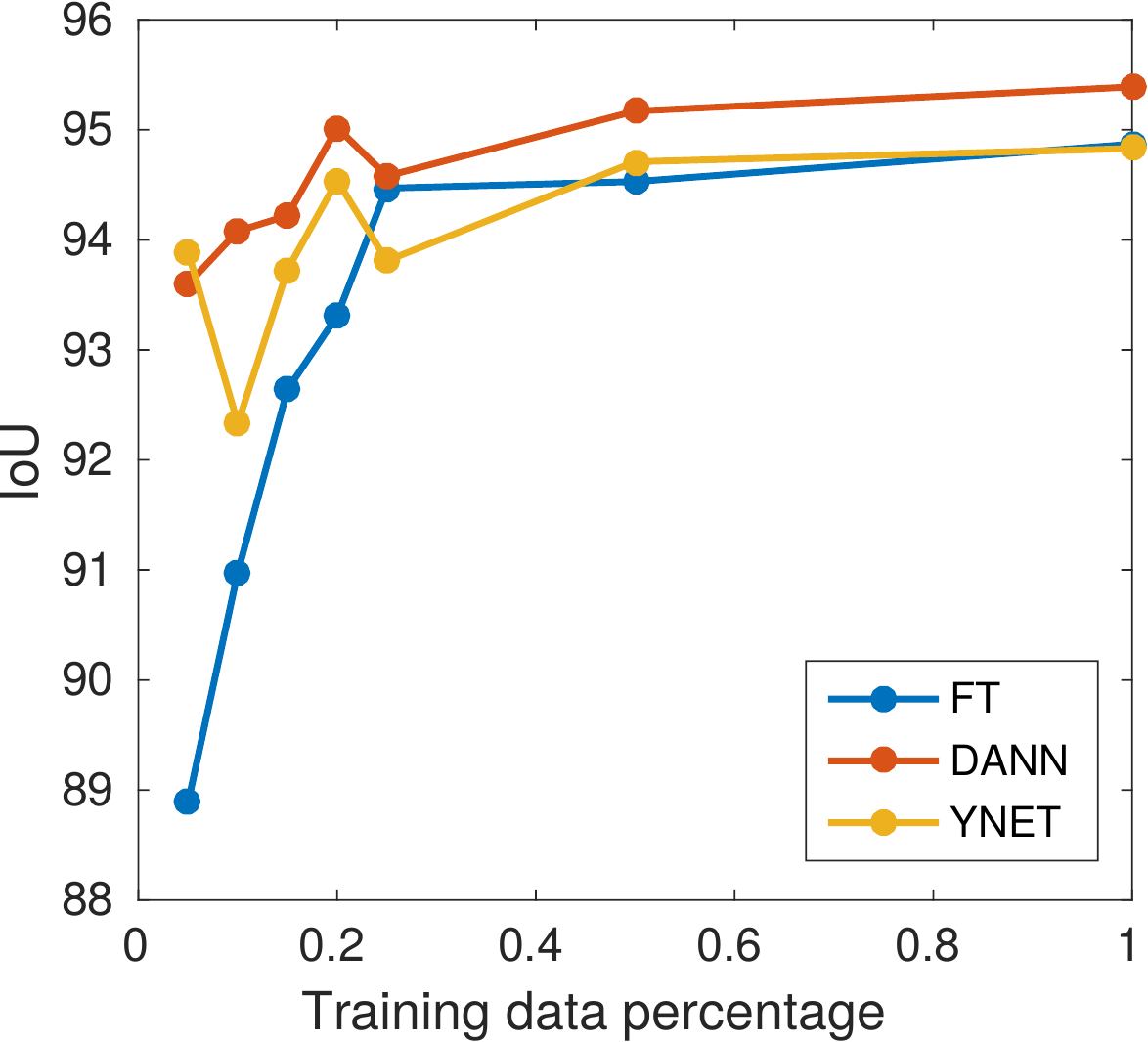} 
    \includegraphics[width=0.48\linewidth]{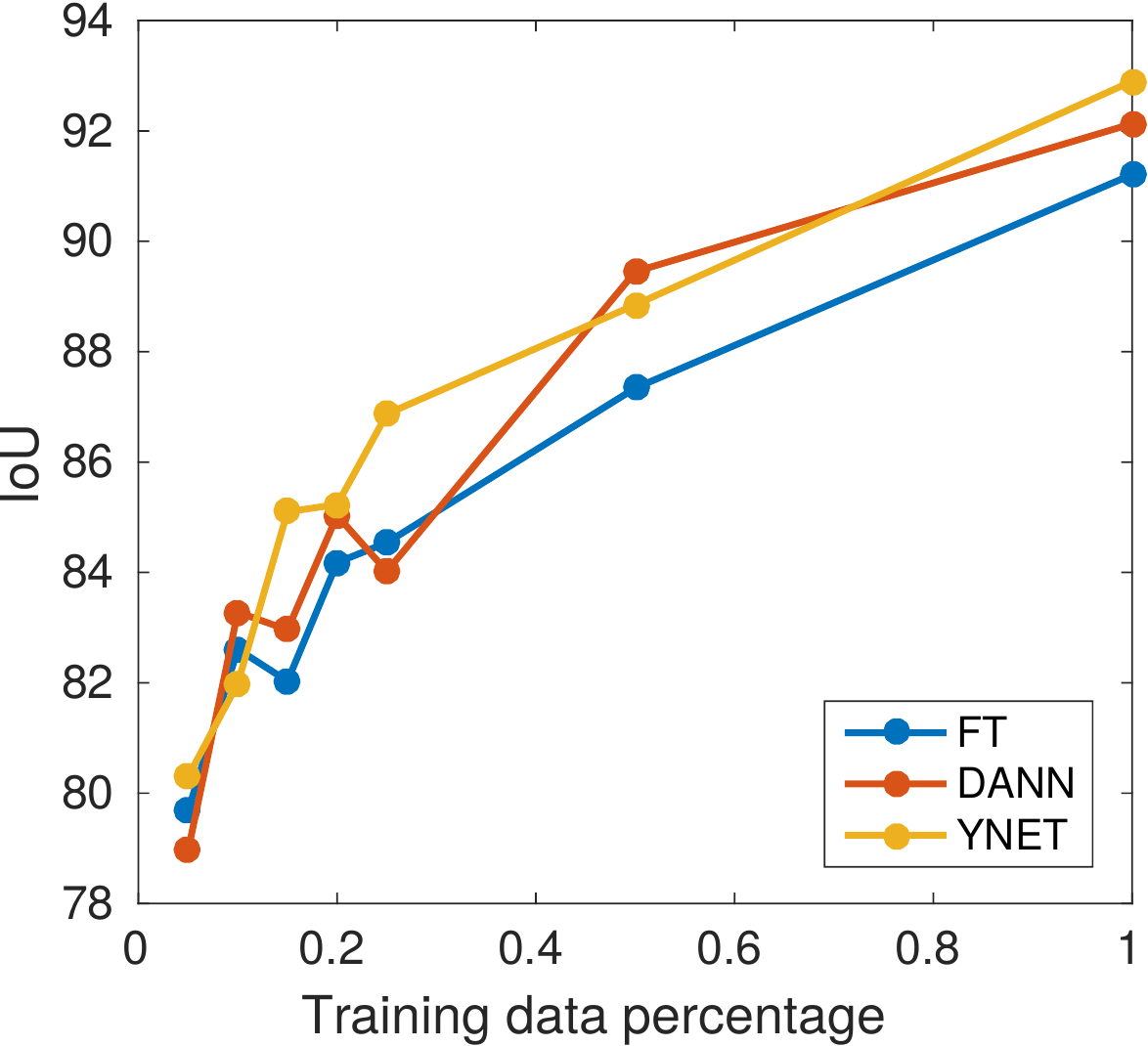} 
	\caption{Segmentation performance on the HeLa (left) and Drosphila dataset (right) after finetuning on various fractions of the target data. }
\label{fig:results-finetuning}
\end{figure}

\begin{figure*}[t!]
	\centering
	\begin{minipage}{0.13\linewidth}
		\centering
		\centerline{Ground truth}
	\end{minipage}
	\begin{minipage}{0.13\linewidth}
		\centering
		\centerline{FT (0\%)}
	\end{minipage}
	\begin{minipage}{0.13\linewidth}
		\centering
		\centerline{DANN (0\%)}
	\end{minipage}
	\begin{minipage}{0.13\linewidth}
		\centering
		\centerline{Y-NET (0\%)}
	\end{minipage}
	\begin{minipage}{0.13\linewidth}
		\centering
		\centerline{FT (15\%)}
	\end{minipage}
	\begin{minipage}{0.13\linewidth}
		\centering
		\centerline{DANN (15\%)}
	\end{minipage}
	\begin{minipage}{0.13\linewidth}
		\centering
		\centerline{Y-NET (15\%)}
	\end{minipage}
	\begin{minipage}{0.13\linewidth}
		\centering
		\includegraphics[width=\textwidth]{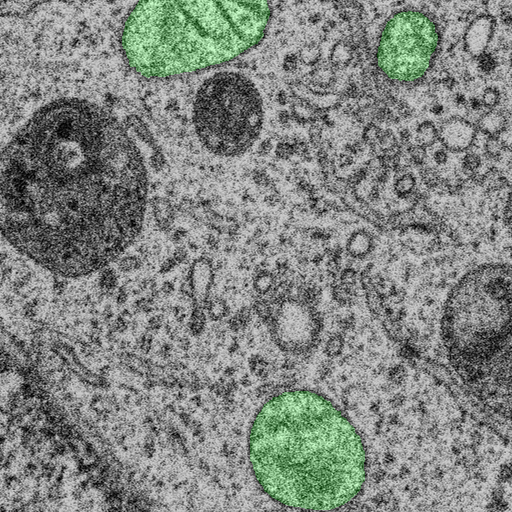}
	\end{minipage}
	\begin{minipage}{0.13\linewidth}
		\centering
		\includegraphics[width=\textwidth]{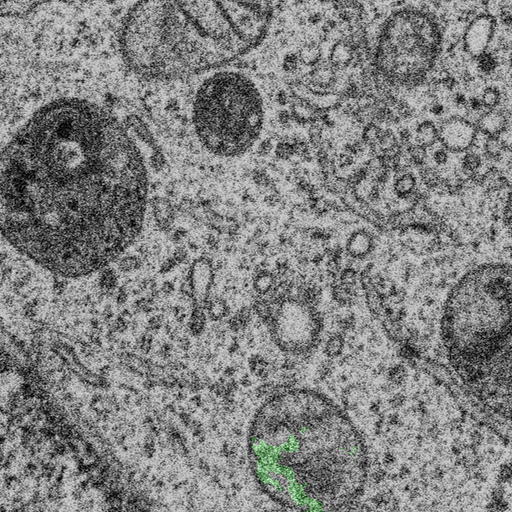}
	\end{minipage}
	\begin{minipage}{0.13\linewidth}
		\centering
		\includegraphics[width=\textwidth]{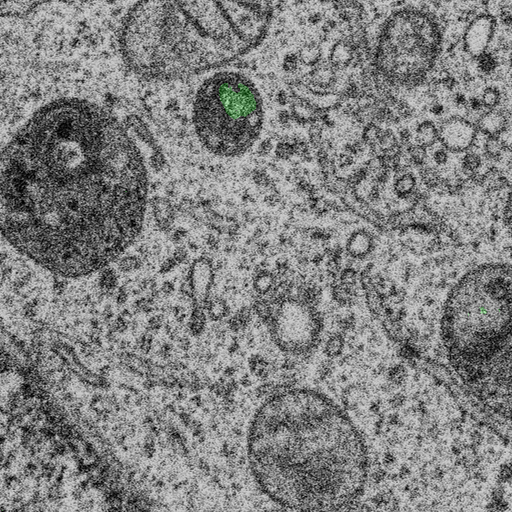}
	\end{minipage}
	\begin{minipage}{0.13\linewidth}
		\centering
		\includegraphics[width=\textwidth]{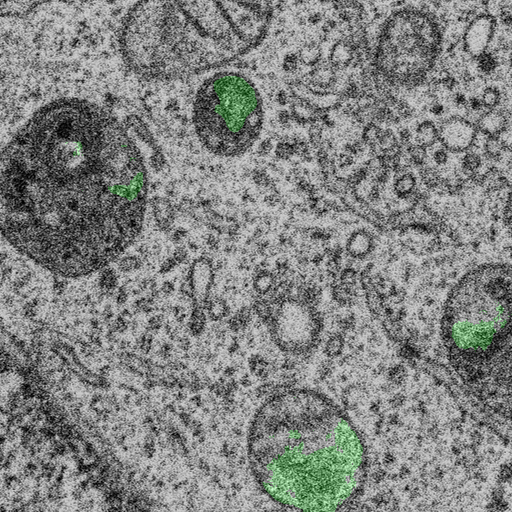}
	\end{minipage}
	\begin{minipage}{0.13\linewidth}
		\centering
		\includegraphics[width=\textwidth]{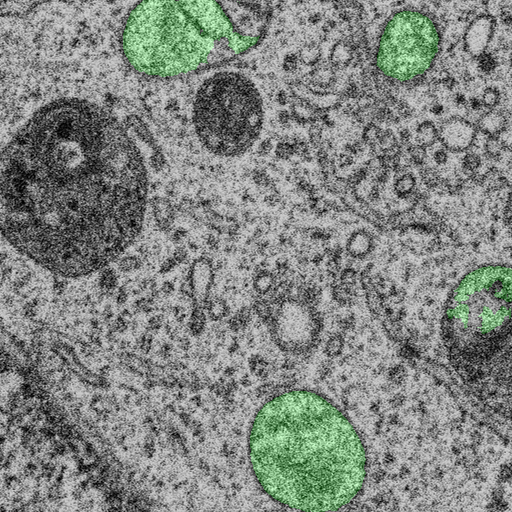}
	\end{minipage}
	\begin{minipage}{0.13\linewidth}
		\centering
		\includegraphics[width=\textwidth]{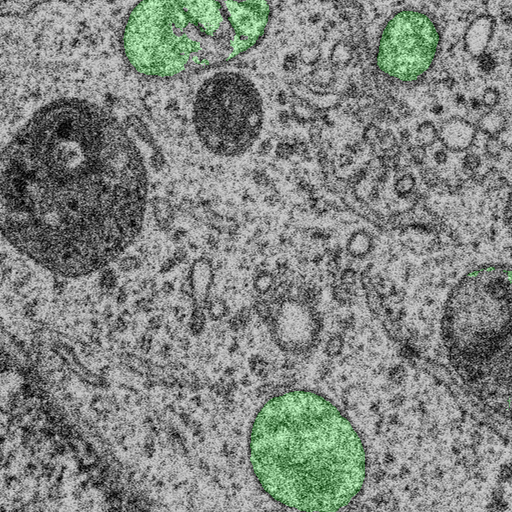}
	\end{minipage}
	\begin{minipage}{0.13\linewidth}
		\centering
		\includegraphics[width=\textwidth]{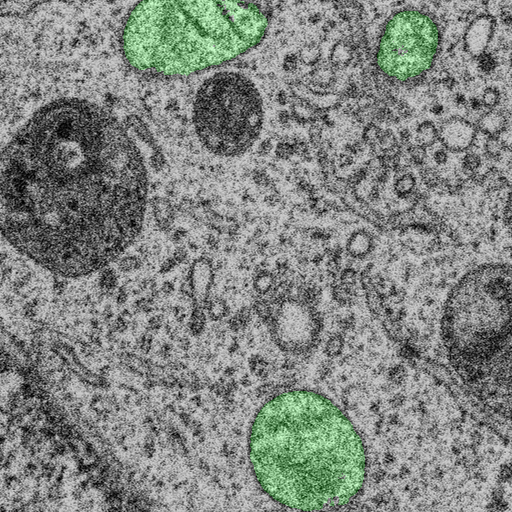}
	\end{minipage}
	\begin{minipage}{0.13\linewidth}
		\centering
		\includegraphics[width=\textwidth]{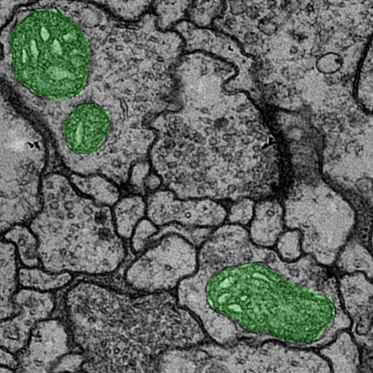}
	\end{minipage}
	\begin{minipage}{0.13\linewidth}
		\centering
		\includegraphics[width=\textwidth]{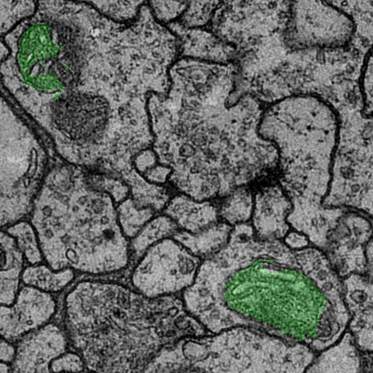}
	\end{minipage}
	\begin{minipage}{0.13\linewidth}
		\centering
		\includegraphics[width=\textwidth]{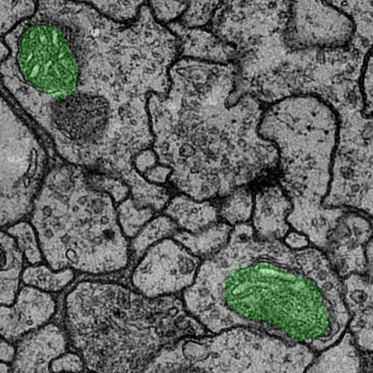}
	\end{minipage}
	\begin{minipage}{0.13\linewidth}
		\centering
		\includegraphics[width=\textwidth]{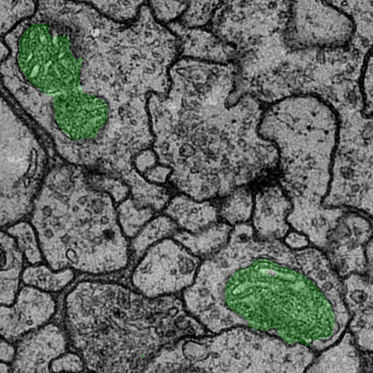}
	\end{minipage}
	\begin{minipage}{0.13\linewidth}
		\centering
		\includegraphics[width=\textwidth]{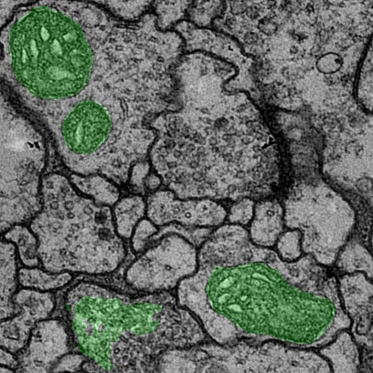}
	\end{minipage}
	\begin{minipage}{0.13\linewidth}
		\centering
		\includegraphics[width=\textwidth]{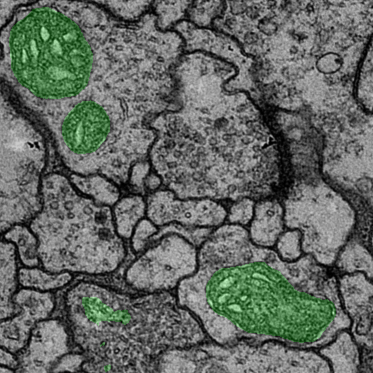}
	\end{minipage}
	\begin{minipage}{0.13\linewidth}
		\centering
		\includegraphics[width=\textwidth]{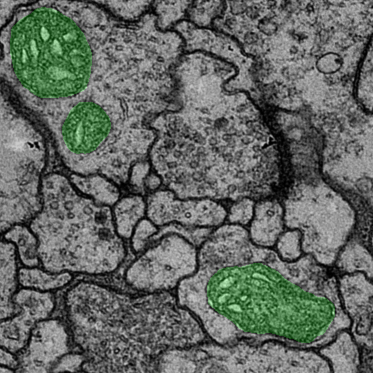}
	\end{minipage}
	\caption{Qualitative comparison of the finetuning baseline (FT), DANN and Y-NET for the HeLa (top) and Drosophila (bottom) dataset. We illustrate segmentation results in the unsupervised and semi-supervised setting (using $15\%$ of the target labels). \commentresolved{insert segmentation results} }\label{fig:qualitative-results}
\label{fig:qualitative-results}
\end{figure*}

\begin{table}[h!]
\centering
\begin{tabular}{|L{0.4cm}|L{0.8cm}|L{1cm}|L{1.2cm}|L{1.1cm}|L{1.05cm}|}
\hline
     & FT    & MMD   & CORAL & DANN  & Y-NET  \\ \hline
H & 8.83 & 2.33 & 3.46 & 11.80 & \bf 22.51 \\ \hline
D & 28.70  & 44.96 & 40.28 & \bf 49.90 & 49.55 \\ \hline
\end{tabular}
\caption{Segmentation performance (in terms of IoU) of the discussed DA approaches on the HeLa (H) and Drosophila (D) dataset in the unsupervised setting. \commentresolved{this will be a graph evaluated for 0\%, 5\%, \dots, 25\%, 50\% and 100\%}}
\label{tab:unsupervised-results}
\end{table}

We compare the methods described in sections \ref{sec:domain-adaptive-segmentation} and \ref{sec:y-net} to the classical finetuning baseline (FT) which pre-trains the segmentation network on the source and finetunes on the available target labels. Segmentation performance on the target test set is measured by means of the intersection-over-union (IoU). \Figref{tab:unsupervised-results} summarizes the unsupervised results for the HeLa and Drosophila dataset. Generally speaking, all the DA approaches significantly outperform the finetuning baseline on the Drosophila data, whereas the domain shift in the HeLa data is too large for MMD and CORAL regularization. For both datasets, DANN is the best performing regularization-based technique. The proposed Y-NET achieves similar to better performance. By finetuning on a fraction of the target labels, we denote that DANN and Y-NET generally outperform the finetuning baseline (\figref{fig:results-finetuning}). \Figref{fig:qualitative-results} shows qualitative segmentation results on the HeLa and Drosophila datasets. Both DANN and Y-NET are able to detect large fractions of mitochondria and outperform the finetuning baseline significantly. Note that the Y-NET approach avoids erroneous detections obtained by finetuning, \eg the upper left mitochondria and the lower left structure in the Drosophila data. 

\commentresolved{discuss the numbers and segmentation results in more detail once they are here.}

\section{Conclusion}
\label{sec:conclusion}

Convolutional neural networks deliver state-of-the-art segmentation results, with the down-side of requiring large amounts of labeled data. Similar shortcomings can be found in all supervised deep learning tasks, but image classification problems have been the target of most domain adaptation work so far. We have demonstrated how the domain adaptation techniques originally proposed for classification can be extend to encoder-decoder segmentation networks. We have also introduced a new DA approach which overcomes the domain shift by training an additional decoder unsupervised on both source and target domains. We believe that the conceptually simple auto-encoding alignment approach will ease the application of CNN-based segmentation in biomedical imaging. In future work, we plan to address unsupervised approaches such as zero-shot learning for segmentation in volume electron microscopy. 

\small
\bibliographystyle{IEEEbib}
\bibliography{main}

\end{document}